%% file: main.tex
\documentclass{ar-1col-S2O}
\usepackage{gb4e}
\usepackage{natbib}
\setcitestyle{authoryear,round}
\usepackage{hyperref}
\usepackage{url}
\usepackage{graphicx}
\jname{Accepted for publication in the Annual Review of Linguistics}
\jvol{13}
\jyear{2027}
\doi{10.1146/annurev-linguistics-040125-043013}

\firstpagenote{*These authors contributed equally to this article}

\begin{document}

% Page header
\markboth{Mercanoglu Sincan et al.}{Machine Translation for Sign Languages}

% Title
\title{Machine Translation for Sign Languages}

%Authors, affiliations address.
%\author{Author B. Authorone,$^1$ Firstname C. Authortwo,$^2$ and D. Name Authorthree$^3$
%\affil{$^1$Department/Institute, University, City, Country, Postal code; email: author@email.edu}
%\affil{$^2$Department/Institute, University, City, Country, Postal code}
%\affil{$^3$Department/Institute, University, City, Country, Postal code}}
\author{Ozge Mercanoglu Sincan, Anton Pelykh$^*$, Edward Fish$^*$, Harry Walsh$^*$, JianHe Low$^*$, Karahan Sahin$^*$, Oline Ranum$^*$, Sobhan Asasi$^*$, Steven Emery$^*$, and Richard Bowden
\affil{Centre for Vision, Speech and Signal Processing (CVSSP), \\University of Surrey, Guildford, UK, GU2 7XH. \\
email: o.mercanoglusincan@surrey.ac.uk}
}

%Abstract
\begin{abstract}
Sign language machine translation has progressed substantially over the past decade, evolving from isolated sign recognition to end-to-end translation systems. Advances in pose estimation, transformer architectures, and large-scale dataset collection have driven progress, yet challenges remain. Datasets are limited compared to spoken-language resources; evaluation metrics inadequately capture the linguistic quality of output; and models must capture the simultaneous, multi-layered, and three-dimensional structure of sign languages. This manuscript provides a comprehensive review that seeks to balance technical challenges with stakeholder considerations. We examine the linguistic properties that make sign languages computationally unique, trace the evolution of recognition, translation, and production systems, and analyze ongoing technical challenges. Crucially, we address ethical considerations around data governance, community involvement, and appropriate use. Drawing on interdisciplinary perspectives spanning computer vision, sign language linguistics, and deaf studies, our analysis emphasizes that continued progress requires sustained collaboration across these fields and with deaf communities.

\end{abstract}

%Keywords, etc.
\maketitle

%Table of Contents
%\tableofcontents

\input{sec/1-intro}
\input{sec/2-linguistics}

% -----------------------------------------------------------
\input{sec/4-ethics}
%-----------------------
\input{sec/3-literature} 
\input{sec/3.1-SLR}

\input{sec/3.3-SLT}

\input{sec/3.4-text2sign}

% -----------------------------------------------------------

\input{sec/5-challenges}
\input{sec/6-conclusion}

% Acknowledgements
\vspace{-10pt}
\section*{ACKNOWLEDGMENTS}
This work was supported by the Innosuisse IICT Flagship (PFFS-21 47), EPSRC grant APP24554 (SignGPT-EP/Z535370/1)  and through funding from Google.org via the AI for  Global Goals scheme. This work reflects only the authors’ views and the funders are not responsible for any use that may be made of the information it contains.
AI tools based on large language models (LLMs) were used by the authors to assist with proofreading and structural suggestions. All recommendations generated by these tools were carefully reviewed and approved by the authors.

\vspace{-10pt}
\section*{DISCLOSURE STATEMENT}
The authors are not aware of any affiliations, memberships, funding, or financial holdings that might be perceived as affecting the objectivity of this review. 
\vspace{-10pt}
\bibliographystyle{ar-style1}   % Annual Reviews style (Surname N.)
\bibliography{main}  % name of .bib file (no .bib extension)

\end{document}

%% file: sec/1-intro.tex
\section{INTRODUCTION}

Machine Translation (MT) aims to develop automated systems that convert meaning across languages. While MT research has historically focused on text and speech, recent advances in computer vision and deep learning have fueled increasing attention to sign languages, which are natural languages that make systematic use of the visual-spatial modality. 
Sign languages differ fundamentally from spoken languages in both form and structure: meaning is expressed through the coordinated use of hands, face, and body \citep{Stokoe1961, Bank2011}; they exploit three-dimensional space to represent grammatical relations and spatial mapping \citep{Baker2016}; and they express multiple layers of linguistic information simultaneously rather than sequentially \citep{Vermeerbergen2007}. These properties make sign languages not only linguistically distinct but also computationally challenging, placing sign language machine translation at the intersection of linguistics, computer vision, natural language processing, and human-computer interaction.

Historically, sign language technologies emerged from efforts to recognize isolated signs using specialized hardware, such as data gloves or motion capture systems \citep{fels1990building, takahashi1991hand}, combined with handcrafted features and rule-based models. Subsequent work moved toward vision-based approaches with statistical sequence models, enabling continuous sign language recognition from video \citep{starner1997wearable, vogler1999parallel}. However, the more recent adoption of deep learning has transformed the field: convolutional and pose-based models have improved visual feature extraction, while transformer architectures have enabled temporal modeling. A major milestone was the move from recognition to translation by formulating sign-to-text as an end-to-end neural task \citep{camgoz2018neural}, and end-to-end pipelines have been developed for both \textit{Sign Language Translation (SLT; sign-to-text)} \citep{gfslt-vlp, signllms} and \textit{Sign Language Production (SLP; text-to-sign)}\footnote{This naming convention follows the computer vision literature. We retain these terms for consistency, using sign language machine translation as the overarching term for both directions.} \citep{saunders2020progressive, zuo2025signs} using animated or photorealistic avatars or digital humans. 

Despite these advances, sign language machine translation performance remains limited. Much of the early work in the field has faced strong criticism from the deaf\footnote{Following recent convention, we use lowercase \textit{deaf} throughout 
to refer to members of the sign language community, without 
distinguishing between culturally Deaf and audiologically 
deaf~\citep{kusters2017innovations}.} community as it has demonstrated a lack of linguistic understanding, engagement with, and sensitivity to stakeholder communities, often trivializing the problem. Reported gains are often tightly coupled to specific datasets, domains, and recording conditions \citep{sincan2025gloss}, while many core linguistic properties of sign languages--such as productive morphology, spatial reference, and non-manual grammatical marking--are only partially modeled or evaluated. Furthermore, the field suffers from a ``data bottleneck". Unlike spoken-language MT, which benefits from vast amounts of web text, sign language datasets are scarce, expensive to collect, and difficult to annotate.
Large-scale resources frequently rely on broadcast interpretations rather than spontaneous, community-based signing, introducing biases in language use, and access to video data has ethical implications \citep{de2025deaf}. As a result, dataset scale, annotation practices, and evaluation protocols profoundly shape what current systems can learn and what they overlook.

Prior surveys have reviewed sign language recognition and 
translation from technical~\citep{rastgoo2021sign, 
DeCoster2024} and interdisciplinary 
perspectives~\citep{bragg2019sign}, but the field has 
evolved substantially since, with the emergence of 
gloss-free translation, LLM-based pipelines, and 
diffusion-based production. This manuscript provides a comprehensive and critical overview of machine translation for sign languages from technical, social, and linguistic perspectives. Rather than focusing solely on architectural advances or benchmark performance, we situate computational approaches within the linguistic structure of sign languages and the social context in which these technologies are developed and deployed. As a hearing and deaf team with varying levels of sign language proficiency across different national sign languages, who work closely with deaf communities in our research, we aim to reflect these perspectives throughout. We begin by outlining linguistic foundations that are essential for understanding sign language processing (Section 2), before establishing the ethical framework that should guide research and development (Section 3). We then trace the historical development of sign language technologies, examining datasets, recognition systems, and methods for both SLT and SLP (Section 4). Section 5 analyzes ongoing technical challenges related to data availability, evaluation, generalization, and computational constraints. We conclude by outlining future research directions that align with technical innovation (Section 6). 

%% file: sec/2-linguistics.tex
\section{LINGUISTIC FOUNDATIONS OF SIGN LANGUAGES}
\label{section2}
Understanding the linguistic structure of sign languages is essential for developing effective computational models. Due to space limitations, this overview is non-exhaustive, but captures some of the key properties relevant to sign language machine translation.

\textbf{Multimodality and Simultaneity:} Sign languages are articulated through manual parameters (handshape, place of articulation, movement and orientations \citep{Stokoe1961, Battison1978}) and non-manual parameters (including facial expressions, head and body movements, eye gaze, mouthings and mouth gestures \citep{Fox2023, Bank2011}).  This multi-channel organization allows morphological, syntactic, and semantic information to be expressed in parallel, supporting a higher degree of simultaneity than the linear structure of spoken languages \citep{Vermeerbergen2007, Baker2016}. For machine translation, this means that models must synchronize visual cues from the hands, face, and torso. A change in a single parameter can alter both lexical and grammatical meaning. Many sign languages mark clause type through non-manual cues, such as raising the eyebrows to indicate a yes/no question, while negation may be expressed through head shakes even when a manual negation sign is absent \citep{Baker2016, DeCoster2024}.

\textbf{Sign Classes:} Signs are organized into distinct classes that vary in form, function, and degree of conventionalization. \textit{Lexical signs} constitute the most frequent class \citep{Fox2023}, which includes lexical content such as nouns, verbs, adjectives, and adverbs \citep{Baker2016}. However, they may account for less than 60\% of discourse \citep{Fenlon2014}.  To express new concepts or proper nouns, 
signers can use \textit{fingerspelling}: language-specific manual alphabets which represent letters from spoken language \citep{DeCoster2024}.  More challenging for computational systems are the \textit{non-lexicalized} or \textit{partially-lexicalized signs}, such as indexing signs, which establish and maintain reference through pointing \citep{Herreweghe2013}, and depicting signs, which use visually grounded outlines of shapes, sizes, and trajectories to represent actions, entities, or relations \citep{Fox2023}. The prevalence of non-lexicalized signing creates an ``open-vocabulary" problem. Recognition systems based on fixed ``gloss\footnote{Gloss is a written representation of a sign using words from spoken language.}" dictionaries are insufficient for productive language.

\textbf{Spatial Grammar and Discourse Structure:}
Signers use both their own body and the physical space around them, referred to as the \textit{signing space}, to construct utterances and structure information. Sign phrases may combine several classes, strategies and channels, shifting between established forms and productive language generation. This makes comprehension heavily context-dependent: systems must discern, for example, whether a motion trajectory is constitutive of a lexical sign or serves a grammatical function such as indicating agreement or spatial depiction.

Sign phrase construction differs structurally from spoken languages. While sign languages also build on syntactic constituents that cannot be split, their ordering is less dependent on linearity and is relatively flexible in most sign languages \citep{Baker2016}. Consequently, dependency relations may differ: a single sign language phrase can map onto multiple spoken-language word orders, and vice versa. For instance, adverbial modification in sign language is often achieved by modifying the manual parameters of a verb (e.g., slowing the movement to indicate ``slowly") rather than adding a distinct adverb sign.

Many sign languages also employ classifier constructions, a productive morphological process where the handshape is selected to reflect properties of the referent \citep{Schembri2003, Baker2016}. For example, a flat handshape might represent a vehicle, and moving the flat hand through space can indicate a car driving or parking. Because the meaning emerges from the combination of handshape and movement rather than from a fixed lexical entry, classifier constructions pose significant challenges for recognition systems that rely on template matching or closed vocabularies.
Non-manual marking also plays a central role at the discourse level. A prominent example is role shift, where structured changes in body posture, eye gaze and facial expressions are used to indicate shifts in discourse perspective.

%% file: sec/4-ethics.tex
\vspace{-10pt}
\section{ETHICS and SAFETY}
\label{sec:ethics}

The development and deployment of sign language technologies have consequences for sign languages, deaf communities, access to services, and the interpreting profession. Recent work across deaf studies, linguistics, and AI ethics has increasingly emphasized that ethical considerations must be embedded from the outset, not treated as an afterthought \citep{Fox2023, desai2024systemic, atwell2025nothing, de2025deaf}. This section synthesizes principles drawn from deaf communities, researchers, and policy documents.

\textbf{``Nothing About Us Without Us":} Across policy documents on AI and sign language ethics, one message is consistent: “nothing about us without us” \citep{charlton1998nothing, atwell2025nothing}. This principle is not intended to be a slogan, but as a concrete governance structure. Research suggests that meaningful inclusion requires deaf people to be in real decision-making roles \citep{Fox2023, desai2024systemic, atwell2025nothing}, not just as “consulted users” at or near the end of a project, but as principal investigators, co-investigators, advisory-board members, or paid consultants shaping methods, testing, and deployment from the outset. 
Projects should be grounded in deaf-led research cultures, where deaf and hearing researchers work together, with deaf cultural norms consciously integrated into every process. 
Furthermore, a diversity of deaf voices is crucial, given that deaf people do not constitute a homogenous socio-linguistic-cultural group. Efforts need to be made to ensure that governance includes Black and Global Majority deaf people, deafblind signers, LGBTQIA+ signers, and people from different regions and social backgrounds, rather than only prominent individuals or representatives from deaf organizations. 

\textbf{Data Governance and Consent:} Large-scale sign language datasets depend on huge volumes of video data of people signing. That raises significant concerns, as sign language data is both \textit{biometric} (including faces, bodies, and gaze) and \textit{cultural}. 
The literature identifies several requirements \citep{crasborn2010does, WFD, bragg2021fate}: (i) accessible, informed consent provided in a native sign language, not just written consent forms; (ii) clear boundaries on data reuse, particularly for open-source release; (iii) explicit Name-image-likeness (NIL) rights policies stating whether and under what conditions someone’s face, body and signing style may be used to create a photorealistic avatar; and (iv) fair remuneration and benefit-sharing mechanisms, ensuring deaf contributors are not treated as unpaid ``raw material" when their data contributes to commercial products or major funding outcomes.

\textbf{Linguistic Integrity and Diversity:} Sign languages are full, complex languages with their own grammar and regional varieties. However, AI systems tend to flatten this richness, drifting toward spoken-language-influenced forms, or producing output that does not consistently reflect any specific sign language.
Avatars or photorealistic synthetic signers risk reflecting how non-signers imagine sign language `looks', particularly if deployed in public-facing tools. If output is simplified or reflects spoken-language influence, it can mislead non-signers and undermine the status of sign languages.  

Sign languages include regional and cultural variations within a country, including minority groups and deafblind practices. There is a risk that such variations may be erased, especially if training data is biased toward certain groups. 
Unlike text-based machine translation, sign language systems must handle depiction, constructed action, and classifiers, which are central to natural signing. AI systems that cannot reliably model these phenomena risk producing unnatural or misleading output. 
This raises questions about linguistic quality standards: what counts as acceptable output for different use-cases, and when should systems indicate that human interpretation is required? A related concern is the provenance of training data. Much of the available large-scale data is derived from broadcast interpretations rather than native deaf signing, which risks embedding spoken-language-influenced forms as representative of natural sign language \citep{Fox2023, de2025deaf}. The technical implications of this are discussed in Section \ref{sec:technical}.

\textbf{Appropriate Use:} A common concern raised by deaf and interpreter communities is that AI should not quietly replace human interpreters, especially in high-stakes settings \citep{de2025deaf}. A consistent position in the literature is that projects should be clear that they aim to support access when human interpreters are unavailable, rather than to displace them \citep{de2025deaf, WFD}. In low-risk contexts such as transport announcements or pre-recorded informational videos, AI-mediated sign language may be appropriate if deaf experts verify quality. However, in high-stakes contexts such as healthcare consultations, legal interviews, and immigration hearings, a core ethical position should be that AI is never an adequate replacement for qualified, registered interpreters. It is also essential to monitor ``access hierarchies" - situations where some deaf people receive high-quality human interpreting while others are offered cheaper AI tools due to cost-saving pressures. Clear boundaries between acceptable and unacceptable use cases should be defined and integrated into documentation and procurement guidance. Several position papers and guidance documents have been produced by organizations such as the World Federation of the Deaf (WFD), British Deaf Association (BDA), Coalition for Sign Language Equity in Technology (COSET), which provide guidelines for the use of AI \citep{WFD, BDA, CoSET}.

\textbf{Governance and Accountability:} 
Responsible deployment requires clear governance structures. This includes transparent documentation of training data, known limitations, and intended use-cases, ideally available in sign language video formats as well as text. When AI tools cause harm or misunderstanding, users should have clear channels for complaints, redress, or opt-out. It is also essential to monitor how AI deployment affects interpreters and deaf professionals, to ensure that these systems do not lead to budget cuts in interpreting services.

\textbf{Summary:} 
Sign language AI sits at the intersection of disability rights, language rights, and digital justice. The decisions that researchers and developers make about leadership, data, language, and use cases will shape how these technologies affect deaf communities. If these decisions are not made carefully, there is a risk of entrenching audism\footnote{Audism, as defined by \cite{humphries1977communicating}, is ``the notion that one is superior based on one’s ability to hear or behave in the
manner of one who hears".} and creating a two-tier system where some deaf people receive quality human interpreting while others are pushed toward inadequate automated tools. But with careful use, AI has the potential to expand access and choice.

%% file: sec/3-literature.tex
\section{HISTORICAL DEVELOPMENT OF SIGN LANGUAGE TECHNOLOGY}
Sign language technology research spans several decades,
with recognition and production often advancing in parallel (Figure 1). The earliest efforts explored sign recognition using image processing \citep{tamura1988}, alongside rule-based translation and synthesis systems \citep{loomis1983, veale1998challenges}, while research in the 1990s adopted approaches from speech recognition \citep{starner1997wearable, vogler1999parallel}. The subsequent adoption of deep learning transformed the field, with \cite{camgoz2018neural} formalizing \textit{sign-to-text translation (SLT)} as an end-to-end sequence-to-sequence task. 
In this section, we trace this evolution across four dimensions: datasets that have enabled computational research (Section \ref{sec:data}), approaches to sign language recognition (Section \ref{sec:slr}), methods for translating signing into text (Section \ref{sec:slt}), and techniques for generating signing from text (Section \ref{sec:slp}). 

\begin{figure}[h]
\includegraphics[width=1\linewidth]{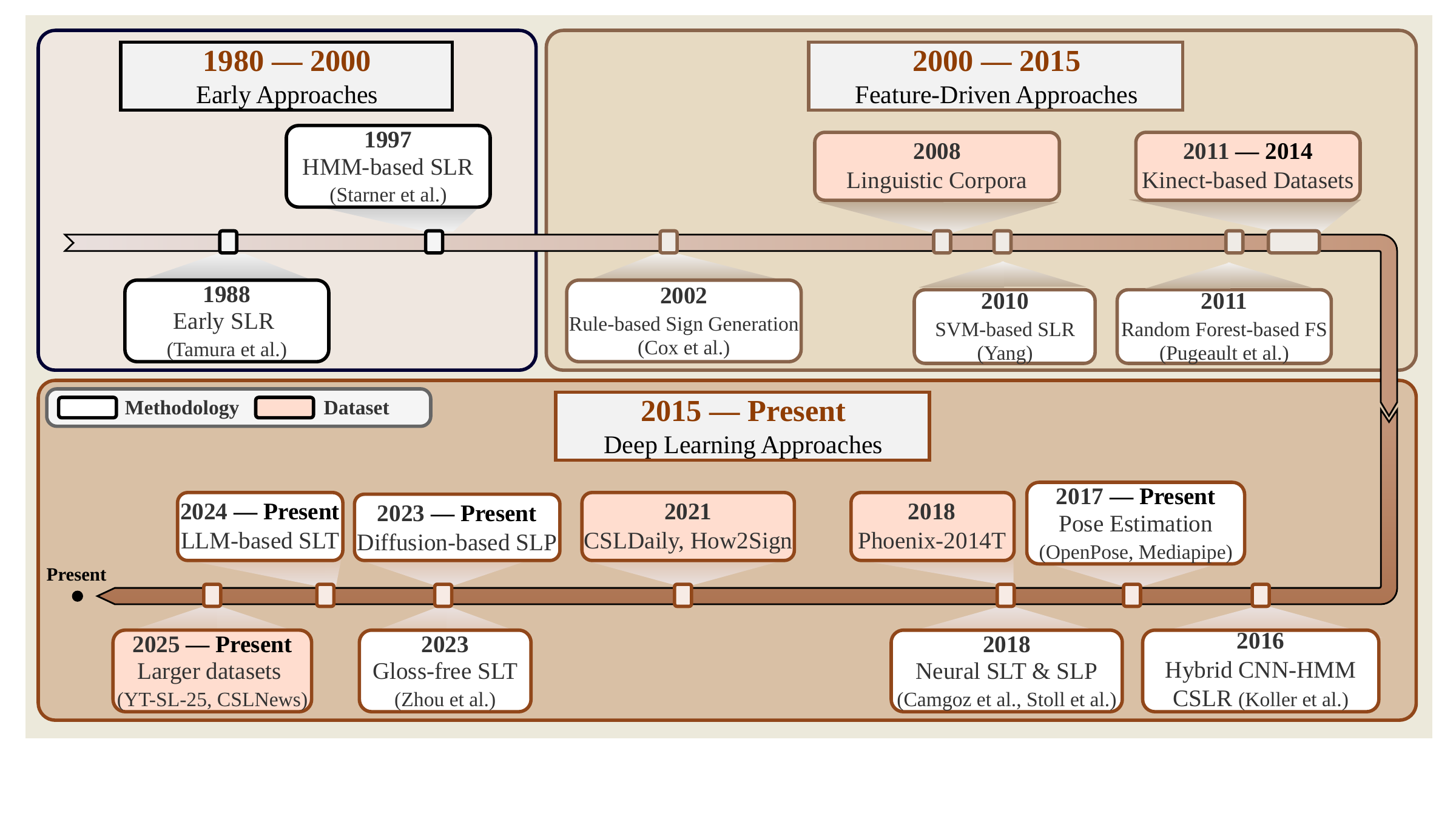}
\caption{A brief history of AI for sign language technologies, highlighting key methodological advances and dataset releases across recognition, translation, and production.}
\label{fig1}
\end{figure}

%% file: sec/3.1-SLR.tex
\vspace{-10pt}
\subsection{Datasets} \label{sec:data}
The development of sign language datasets has been closely tied to available sensing technologies and task formulations. 
Early datasets in the 1990s relied on wearable sensors \citep[e.g.,][]{fels1990building}, such as data gloves, optical markers, and magnetic tracking systems, which enabled precise motion capture but were intrusive, difficult to scale, and unsuitable for real-world deployment. Advances in computer vision gradually shifted dataset collection toward vision-based approaches \citep{starner1997wearable}. Later work used RGB cameras and depth sensors (e.g., Microsoft Kinect) to enable non-intrusive recording while facilitating simple geometric and skeletal extraction. More recently, monocular RGB video has become the dominant modality due to its ubiquity and compatibility with deep learning methods. Table~\ref{tab:dataset} provides an overview of major datasets. For a more detailed comparison of scale, benchmark performance, and research adoption, we refer the reader to \cite{sincan2025gloss}.

\textbf{Isolated Datasets:}  
Early vision-based datasets were predominantly designed for isolated sign recognition with limited vocabularies or low signer diversity \citep{martinez2002purdue, athitsos2008american}.
From the early 2010s onwards, the dataset scale began to expand in terms of vocabulary size, number of signers, and sample count, such as DEVISIGN \citep{chai2014devisign}, MS-ASL \citep{joze2018ms}, WLASL \citep{li2020word}, ASL Citizen \citep{desai2023asl}, AUTSL \citep{sincan2020autsl}, BosphorusSign22K \citep{ozdemir2020bosphorussign22k}, and PopSign-ASL \citep{popsign2023}. These datasets expanded vocabulary coverage to thousands of signs, increased signer diversity, and introduced in-the-wild recording conditions including web-sourced and smartphone-collected data.
In parallel, ChicagoFSWild \citep{shi2018american} provides fingerspelling instances from in-the-wild video.

\begin{table*}
	\centering
    	\caption{
	Overview of sign language datasets. Datasets are characterized by language (Lang), video type (isolated or continuous signing), recording source, number of signers, gloss vocabulary, text vocabulary size (Vocab), number of sentences (Sent), and duration in hours. N/A: annotations not available; `-': unreported statistics.
	}
	\label{tab:dataset}
	\resizebox{1\linewidth}{!}{
	\begin{tabular}{lccccccccc}
		\hline
		\textbf{Dataset} & \textbf{Lang} &
		\textbf{Type} & 
		\textbf{Source} & 
		\textbf{Signers} & \textbf{Gloss} & \textbf{Vocab} &
		\textbf{Sent} & \textbf{Hours}  \\
		\hline
		
        Purdue RVL-SLLL \citep{martinez2002purdue}
        & ASL
		& Isolated 
		& Lab
	%	& Open
		& 14 & 39 & N/A & N/A & -  \\

        ASLLVD \citep{athitsos2008american}
        & ASL
		& Isolated 
		& Lab
	%	& -
		& 6 & 2742 & N/A & N/A & 4  \\

        Corpus NGT \citep{crasborn2008corpus}
        & NGT
		& Continuous 
		& Lab
		& 100 & - & - & 15K & 72 \\

		BSL Corpus \citep{schembri2013building}
        & BSL
		& Continuous 
		& Lab
	%	& Conversational
		& 249 & $\sim$1800 & - & - & 125  \\

		DGS Corpus \citep{konrad2020meine}
        & DGS
		& Continuous
		& Lab
	%	& Conversational
		& 330 & $\sim$10K & 18K & - & 50 \\

        SIGNUM \citep{agris2010}
        & DGS
		& Continuous 
		& Lab
	%	& -
		& 25 & 450 & 1K & 780 & 55   \\

		DGS Kinect 40 \citep{cooper2012sign}
        & DGS
		& Isolated
		& Lab
	%	& -
		& 15 & 40 & N/A & N/A & - \\

            RWTH-PHOENIX-Weather \citep{forster-etal-2012-rwth}
            & DGS
		& Continuous
		& TV
%		& Weather
		& 7 & 911 & 1489 & 2640 & 3  \\

        DEVISIGN \citep{chai2014devisign}
            & CSL
		& Isolated
		& Lab
	%	& Open
		& 8 & 2000 & N/A & N/A & -  \\

		MS-ASL \citep{joze2018ms}
        & ASL
		& Isolated
		& Web
	%	& Open
		& 222 & 1000 & N/A  & N/A & 25  \\

		Phoenix2014T \citep{camgoz2018neural}
        & DGS
		& Continuous
		& TV
%		& Weather
		& 9 & 1066 & 2,887 & 8K & 11  \\

		SMILE \citep{ebling-etal-2018-smile}
        & DSGS
		& Isolated 
		& Lab
%		& Open
		& 40 & 100 & N/A & N/A & - \\
        
		KETI \citep{ko2019neural}
        & KSL
		& Continuous
		& Lab
%		& Open
		& 14 & 524 & - & 104 & 27 \\
                
		BosphorusSign22K \citep{ozdemir2020bosphorussign22k}
        & TID
		& Isolated
		& Lab
%		& Open
		& 6 & 744 & N/A & N/A & 19 \\

		AUTSL \citep{sincan2020autsl}
        & TID
		& Isolated 
		& In/outdoor
%		& Open
		& 43 & 226 & N/A & N/A & 21 \\

        WLASL \citep{li2020word}
        & ASL
		& Isolated
		& Lab
%		& Open
		& 119 & 2000 & N/A  & N/A  & 14  \\
		
        GSL \citep{adaloglou2021comprehensive}
        & GSL
		& Continuous 
		& Lab
%		& Open
		& 7 & 310 & - & 331 & 9 \\

        CSL-Daily \citep{zhou2021improving}
        & CSL
		& Continuous 
		& Lab
%            & Daily
		& 10 & 2000 & 2K & 21K & 23  \\

		BOBSL \citep{albanie2021bobsl}
        & BSL
		& Continuous
		& TV
%		& Diverse
		& 39 & N/A & 78K & 1.2M & 1467 \\

		How2Sign \citep{duarte2021how2sign}
        & ASL
		& Continuous
		& Lab
	%	& Instructional
		& 11 & N/A & 16K & 35K & 79  \\

		OpenASL \citep{shi2022open}
        & ASL
		& Continuous 
		& Web
%		& Open
		& 220 & N/A & 33K & 98K & 288 \\

		ASL Citizen \citep{desai2023asl}
        & ASL
		& Isolated 
		& Web
%		& Open
		& 52 & 2731 & N/A & N/A & - \\

		PopSign v1.0 \citep{popsign2023}
		& ASL
		& Isolated 
		& Phone
%		& Open
		& 47 & 250 & N/A & N/A & - \\
        
		Youtube-ASL \citep{uthus2023youtube}
		& ASL
		& Continuous 
		& Web
%		& Open
		& 2519 & N/A & 60K & 610K & 984 \\

        AfriSign \citep{takyi2025afrisign}
        & Multi
		& Continuous 
		& Web
%		& Open
		& 51 & N/A & 20K & 29K & 60 \\

		CSL News \citep{li2025unisign}
		& CSL
		& Continuous 
		& TV
%		& Diverse
		& N/A & N/A & 5K & 751K & 1985  \\

        YouTube-SL-25 \citep{tanzer2024youtubesl25}
        & Multi
		& Continuous 
		& Web
%		& Open
		& 3000 & N/A & - & 2.16M & 3207  \\
		\hline
	\end{tabular}
	}
\\[0.5ex]
\begin{minipage}{\linewidth}
\scriptsize 
\textit{Abbreviations:} ASL, American Sign Language; BSL, British Sign Language; CSL, Chinese Sign Language; DGS, German Sign Language; DSGS, Swiss-German Sign Language; GSL, Greek Sign Language; KSL, Korean Sign Language; NGT, Sign Language of the Netherlands; TID, Turkish Sign Language.
\end{minipage}
\end{table*}

\textbf{Continuous Datasets:}  Continuous datasets contain unsegmented signing that typically requires human annotation, e.g., sign boundaries and sentence-level alignment. Early examples were limited in scale \citep{agris2010, forster-etal-2012-rwth}.
RWTH-PHOENIX-Weather-T (Phoenix14T) \citep{camgoz2018neural} addressed this gap and became the standard benchmark for sign language translation research, with approximately 1,000 unique glosses, and 8,000 sequences sourced from broadcast television weather forecasts. 
CSL-Daily \citep{zhou2021improving}  expanded coverage to daily life vocabulary with over 2,000 glosses, 21,000 sequences from 10 signers, collected in controlled environments. 

Because annotation is costly and labor-intensive, more recent datasets have been collected with sentence-level alignments only, employing gloss-free approaches (Section \ref{sec:slt}) that overcome the requirement for gloss-level supervision. How2Sign \citep{duarte2021how2sign} comprises approximately 35,000 sign sequences from the instructional domain aligned with spoken-language sentences covering a vocabulary of 16,000 words, recorded from 11 signers over 79 hours of studio data.
At a larger scale, BOBSL \citep{albanie2021bobsl} contains 1,467 hours of BBC broadcast content with more than 1.2 million sentences, and CSL-News \citep{li2025unisign} provides 1,985 hours of Chinese sign language from broadcast news. 
In addition to broadcast sources, several datasets have been collected from web platforms. 
OpenASL \citep{shi2022open} introduced approximately 288 hours of ASL data with a vocabulary of over 33,000 items collected from web sources and involving more than 200 signers. YouTube-ASL \citep{uthus2023youtube} further increased both scale and diversity, providing nearly 1,000 hours of footage, a vocabulary exceeding 60,000 items, and contributions from over 2,500 signers. Efforts to expand coverage to more sign languages include AfriSign \citep{takyi2025afrisign},  JWSign \citep{gueuwou2023jwsign}, and YouTube-SL-25 \citep{tanzer2024youtubesl25}, which spans 25 languages over 3,207 hours.
However, even as the largest dataset currently available, it remains orders of magnitude smaller than the resources available for text and audio.

%% -------
Although considerably smaller, in parallel, sign language corpora have been developed for linguistic research, such as the BSL Corpus \citep{schembri2013building}, Corpus NGT \citep{crasborn2008corpus}, and the DGS Corpus \citep{konrad2020meine}, providing sentence-level recordings with rich linguistic annotations from conversational data. 
Lexical resources and sign dictionaries, e.g., DW-DGS \citep{konrad2020meine}, ASL-Lex \citep{caselli2017asl} further support the field.

\vspace{-10pt}
\subsection{Sign Language Recognition} \label{sec:slr}

\textbf{Isolated Sign Language Recognition (ISLR):}
Early SLR systems relied on hand-crafted features classified using Hidden Markov Models (HMMs) \citep{vogler1999parallel}, Dynamic Time Warping (DTW) \citep{jangyodsuk2014sign}, Support Vector Machines (SVM) \citep{yang2010chinese}, or sub-unit based approaches \citep{cooper2012sign}. With the advent of deep learning, SLR systems transitioned to neural architectures, leveraging 2D and 3D Convolutional Neural Networks (CNNs) to extract spatial and spatio-temporal representations \citep{joze2018ms, li2020word}, and more recently to vision and video transformer architectures for temporal modeling \citep{gueuwou-etal-2025-shubert}. 
In parallel, improvements in camera technology (e.g., Microsoft Kinect) and pose estimation frameworks (e.g., OpenPose \citep{cao2017realtime}, Mediapipe \citep{lugaresi2019mediapipe}) have shifted research toward skeletal representations offering interpretable, appearance-invariant modeling, often processed with Graph Neural Networks (GCNs) or integrated into fusion-based architectures to leverage complementary information across modalities \citep{jiang2021skeleton}. 

To alleviate the reliance on limited labeled data, recent strategies have turned pretraining: a two-stage approach where models first learn general representations from large unlabeled or weakly-labeled datasets, and are subsequently fine-tuned on a smaller labeled dataset for a specific task.
%to self-supervised and semi-supervised pretraining. 
Several approaches \citep{hu2023signbert+, zhao2024masa} utilize masked prediction objectives over skeletal sequences to learn contextualized representations. Furthermore, it has been demonstrated that significant gains can be achieved for large-vocabularies by learning robust features through pose-guided or cross-modal supervision \citep{jiang2024signclip, wong2025signrep, wu2025cross}.

\textbf{Continuous Sign Language Recognition (CSLR):}
While continuous sign language recognition shares many visual feature extraction and backbone choices with isolated recognition, it introduces additional challenges due to unsegmented input and sequence-level supervision. 
Continuous sign language recognition is predominantly formulated as sequence-level gloss optimization, where the objective is to predict ordered gloss sequences without explicit temporal boundaries \citep{koller2016deep, pu2019iterative, cui2019deep}.  
Early approaches are trained in fully supervised settings using RNN-based or transformer-based sequence models and rely on Connectionist Temporal Classification (CTC) \citep{cihan2017subunets, hao2021self} to align visual inputs with gloss sequences.
Recent CSLR research has shifted toward larger-scale settings and weaker supervision. \cite{raude2024tale} introduced a large-vocabulary CSLR benchmark alongside a multi-task framework that jointly performs continuous recognition and sentence retrieval by learning a shared sign-text embedding space.
Recent work \citep{guo2025bridging} leverages large language models to produce frame-level pseudo-labels, enabling CSLR training with substantially reduced annotation and signaling a broader move toward hybrid vision-language supervision.

\textbf{Other Recognition Systems:}
Beyond ISLR and CSLR, other recognition tasks have received attention. Fingerspelling recognition is widely studied, focusing on identifying sequences of hand configurations corresponding to characters of a written alphabet, typically used to spell proper nouns, technical terms, or out-of-vocabulary words within sign language discourse \citep{pugeault2011spelling, shi2018american, shi2019fingerspelling, Prajwal22a}. 
Sign spotting or sign retrieval addresses locating target signs within longer videos using various supervision sources such as dictionaries and lip-reading \citep{jiang2021skeleton, albanie2020bsl}. Sign segmentation, which identifies temporal boundaries of individual signs, is commonly treated as a standalone task supporting continuous recognition \citep{renz2021sign, lowhands, moryossef2023linguistically}.

%% file: sec/3.3-SLT.tex
\vspace{-10pt}
\subsection{Sign Language Translation}
\label{sec:slt}
SLT aims to convert continuous sign videos into spoken or written language sequences. 
Approaches can be broadly categorized into \textit{gloss-based} and \textit{gloss-free} methods based on whether they use glosses as an intermediate representation.

\textbf{Gloss-Based Approaches:} Early systems adopted a two-stage pipeline: first, predicting an intermediate gloss sequence, representing sign-level lexical units, before translating it into text~\citep{camgoz2018neural,sltunet, ts-slt,zhou2021improving,chen2022simple,yin2021simulslt}. 
 This decomposition enabled the translation stage to be treated as a text-to-text problem, leveraging established machine translation techniques. However, it also introduces a dependency on expensive and labor-intensive gloss annotations, which limits data scalability and model generalization across languages. Additionally, gloss representations abstract away much of the richness of signing, including non-manual features, non-lexical signs, spatial modulation, and depicting construction, making it difficult for downstream translation models to recover this information.

\textbf{Gloss-free Approaches:} To overcome these challenges, recent works have increasingly adopted \textit{gloss-free} frameworks that directly translate visual features into text, a shift largely enabled by the success of transformer architectures~\citep{gfslt-vlp,sign2gpt,signllms,liang2024llavasltvisuallanguagetuning}.
In this direction, several studies use pretraining strategies based on contrastive learning, which teaches the model to place semantically similar signs and sentences close together in representation space ~\citep{gfslt-vlp,sign2gpt,chen2025c}, with extensions exploring geometric constraints to better capture the hierarchical nature of sign movements \citep{fish2025geo}.

With the advent of large language models (LLMs), more recent approaches leverage pretrained linguistic priors to improve translation fluency and coherence ~\citep{sign2gpt,chen2025c}. 
Other works incorporate LLMs or Multimodal LLMs directly into pretraining, utilizing their reasoning capabilities to extract structured motion and semantic representations ~\citep{guo2025bridging,kim2025leveraging,asasibeyond}. A context-aware framework feeds visual features alongside automatically extracted textual cues, such as captions describing the broadcast context, prior-sentence translations, and pseudo-glosses, to improve translation quality  ~\citep{sincan2023context, jang2025lost}. 

Beyond architectural changes, recent studies explicitly investigate \textit{scaling}.   
~\cite{zhang2024scaling} train a model that learns to translate across multiple sign-to-text language pairs simultaneously, by scaling up the size of the training data, the size of the model, and the number of language pairs.
~\cite{li2025unisign} introduce a large-scale dataset and unify pre-training and fine-tuning under a single objective, combining visual appearance and pose.

%% file: sec/3.4-text2sign.tex
\newcommand{\revise}[1]{\textcolor{red}{#1}}
\newcommand{\blue}[1]{\textcolor{blue}{#1}}
\vspace{-10pt}
\subsection{Sign Language Production}
\label{sec:slp}

The core challenge of SLP is bridging the dimensionality gap between a linear symbolic input and a spatio-temporal visual output.
% Early computer graphics avatars ------------
Early paradigms addressed this complexity through modular two-stage pipelines: text-to-gloss translation followed by gloss-to-pose synthesis. These systems initially relied on rule-based architectures and direct lookup of pre-recorded motion fragments \citep{cox2002tessa, elghoul2011websign, zwitserlood2004synthetic}. However, the naive concatenation of segments results in ``robotic" discontinuities and a severe lack of co-articulation. The subsequent shift toward statistical machine translation methods has allowed patterns to be learned from parallel corpora, removing the need for manual rule-coding. Yet, these models still struggled with the high-dimensional complexity of full-body movement. Moreover, the unrealistic nature of graphical avatars was unpopular with the deaf community \citep{kipp2011assessing, wolfe2021myth}, due to their cartoon-like appearance and lack of expressive nuance \citep{rastgoo2021sign}.

% Early Deep Learning ------------
The advent of deep learning saw a transition in the field toward neural text-to-gloss architectures, replacing heuristic rules with sequence-to-sequence models and substituting graphical avatars with GAN-based video generation models. While early neural approaches \citep{stoll2018sign, saunders2020progressive} maintained the explicit gloss intermediary to isolate task complexity, this structure introduced a significant information bottleneck \citep{muller2023considerations}. 
Conversely, attempts to bypass this bottleneck via direct text-to-pose mapping \citep{zelinka2020neural, saunders2020progressive} frequently encountered the problem of regression to the mean, with models often predicting under-articulated motion. To mitigate this, hybrid architectures \citep{saunders2021mixed, hwang2021non} have proposed using glosses as auxiliary supervisory signals, allowing for richer latent representations while retaining linguistic grounding.

% Motion tokens ------------------
In pursuit of a more powerful representation, the field has shifted toward learning intermediate motion tokens directly from data. These tokens are compact, discrete units that represent recurring patterns of sign motion. The first works to do this utilized Vector Quantized-Variational Autoencoders (VQ-VAE) to compress pose sequences \citep{yu2024signavatars} and video frames \citep{xie2024sign} into such tokens. While effective for capturing overall body movement, these holistic tokens often prioritized large torso movements at the expense of finer details in hand and facial articulation. Subsequent approaches \citep{yin-etal-2024-t2s, wang2025advanced} have sought to improve the token efficiency and expressive range by means of more advanced quantization. Work has also been done \citep{zuo2025signs, xia2026signmask} on decoupling inputs for hand and body, processing them through separate streams to increase the production quality of manual features.

% Diffusion + LLMs -----------------
While the field has explored advanced latent representations for sign language, classic pipelines that employ glosses have been refined by integrating modern techniques to address their inherent limitations. Consequently, diffusion models have proven to be effective for gloss-to-pose mapping \citep{xie2024g2p}, generating smooth transitions between poses \citep{tang2025discrete}, and pose-to-video generation \citep{fang2025signdiff, shi2024pose}. Similarly, LLMs have been used for text-to-gloss \citep{zhang2025towards} and gloss-to-pose conversion \citep{fang2025signllm}. On the other hand, direct text-to-sign diffusion was also explored in the literature to eliminate intermediate representations\citep{qi2024signgen, baltatzis2024neural}.

%% file: sec/5-challenges.tex
\vspace{-10pt}
\section{TECHNICAL CHALLENGES}
\label{sec:technical}

While sign language technologies have advanced considerably, a number of technical challenges limit the performance and usability of these systems in real-world cases. These challenges arise from the linguistic properties of sign languages, the lack of data and accurate annotations, and issues with effective and accurate evaluation. In this section, we describe these key challenges, addressing five key areas: data availability, evaluation methodology, real-world generalization, computational constraints, and linguistic complexity.

\textbf{Data Availability and Annotation:} The field still lacks large-scale, high-quality datasets for sign language machine translation. In contrast to spoken-language machine translation, where datasets often contain tens or hundreds of millions of sentence pairs, existing datasets are orders of magnitude smaller.
There are multiple reasons for the lack of well-annotated data. Sign language datasets are significantly more expensive to collect and annotate, requiring native or near-native signers to produce content and bilingual annotators for labeling. Even with skilled annotators, annotation poses additional challenges \citep{datasets}. Continuous signing lacks clear segmentation cues, and the relationship between signs and words is rarely one-to-one, violating the monotonic alignment assumptions underlying many sequence-to-sequence architectures ~\citep{linearityinsign}.
In addition, a substantial portion of available large-scale datasets are derived from interpreted content, such as broadcast interpretations rather than native signing. While such data enables collection at scale, it introduces additional challenges. Interpreted signing is shaped by time constraints and interpreter strategies, resulting in weak temporal alignment and approximate sentence-level correspondences that complicate both training and evaluation. Furthermore, as noted in Section \ref{sec:ethics}, this reliance on interpreted data also raises concerns regarding how representative the resulting systems are of natural sign language.

\textbf{Evaluation Metrics:} Evaluation metrics for SLT have been largely borrowed from spoken-language machine translation, including BLEU~\citep{bleu}, ROUGE~\citep{rouge}, and more recently learned metrics such as BERTScore~\citep{bertscore} and BLEURT~\citep{BLEURT}. However, BLEU and similar n-gram overlap metrics penalize valid paraphrases and alternative word orders, which is a particular concern for sign-to-text translation given the flexible phrase structure of sign languages. This problem is compounded when languages exhibit rich morphology, as single tokens can carry dense semantic content that n-gram metrics fail to capture, for both sign languages and agglutinative spoken languages \citep{karahan}. Moreover, these metrics treat all translation errors as equivalent, failing to distinguish between errors that preserve meaning and those that fundamentally alter or obscure it. Some recent works have included LLM evaluation metrics where a Language Model is prompted \citep{jang2025lost}, though this raises concerns regarding reproducibility and evaluator bias.

For SLP, the evaluation is even more challenging. There is currently no agreed metric for assessing the linguistic quality, comprehensibility, or naturalness of generated signing. Human evaluation remains the gold standard but is expensive, difficult to standardize, and itself complicated by questions of who constitutes an appropriate evaluator as deaf signers, interpreters, and linguists may apply different criteria and priorities~\citep{rastgoo2021sign}, and there is no established standard for what level of sign language proficiency evaluators should have.
Back-translation approaches~\citep{saunders2020progressive} evaluate generated signing by translating it back into text using a sign language translation model, then comparing this output to the original text. However, this approach has several limitations: errors may originate from either the production or the translation component, outputs that are machine-readable but unnatural to human viewers may be rewarded, and non-manual features or other multimodal aspects of the generated signing cannot be assessed since the evaluation is reduced to text. More recent metrics propose to capture what humans perceive as the important characteristics in sign production~\citep{bowdenSLPTiers}. 

\textbf{Real-World Variability and Generalization:} Most models are trained and evaluated under relatively controlled conditions, with limited variation in signer identity, signing style, or conversational context. 
For example, standard benchmarks such as Phoenix14T  feature professional interpreters in studio environments with consistent conditions, constrained discourse (in this case, weather), and the language used does not reflect conversational signing in everyday deaf communities.  
 Models frequently overfit to the visual appearance, hand shape, and signing style of individuals in the training set.

To avoid overfitting on visual features, recent methods have explored skeleton or 3D mesh representations to decouple motion from appearance. However, they are often constrained by the error margins of estimation algorithms or the high computational cost of mesh extraction, which limits their use in real-time systems.
Visual variability in real-world deployment poses additional obstacles. Background clutter, variable lighting, partial occlusions from clothing or objects, and camera motion can all degrade pose estimation and feature extraction pipelines that were developed and tuned on studio-quality footage. Another weakness in many datasets is the lack of multiple signer interactions and conversational data, which introduces additional challenges for modeling natural discourse.

\textbf{Real-Time Processing and Computational Constraints:} For practical impact, sign language machine translation systems require real-time or near-real-time performance. However, current state-of-the-art models are not able to deliver real-time translation at high quality. Many factors impact inference performance, including visual pre-processing such as pose extraction, active signer identification, background removal, and cropping. Models that utilize LLMs have further associated inference costs and network speed limitations if calling external APIs.
These issues are even more prominent in mobile applications or embedded devices, which impose additional constraints on model size, memory footprint, and power consumption. Knowledge distillation, quantization, and efficient architecture design offer partial solutions, but achieving acceptable quality under these constraints remains difficult. 

\textbf{Linguistic Complexity:} Sign languages exhibit linguistic structures that challenge the assumptions embedded in current machine learning architectures. As discussed in Section \ref{section2}, non-manual features are an essential component of sign languages, but most state-of-the-art approaches do not include expression, eyebrow position, head tilt, or body posture as an explicit modality. 
Productive morphology poses challenges for both directions, especially for systems relying on fixed vocabularies or template matching. 

Spatial indexing and coreference add further complexity \citep{Baker2016}. Sign languages establish referents at locations in the signing space and subsequently refer back to them through pointing, eyegaze, or movement toward those locations. Resolving coreference requires maintaining a spatial map of these discourse referents across an extended context, which acts as a form of grounded and spatially indexed memory that differs fundamentally from the token-level attention mechanisms typical of current sequence models. Indices can appear over long context windows covering an entire conversation, while current SLT methods are constrained to short context windows, due to the high processing and memory cost of the multi-modal data. 

%% file: sec/6-conclusion.tex
\vspace{-10pt}
\section{CONCLUSION}

Sign language machine translation sits at a unique intersection of computer vision, linguistics, and deaf community perspectives. This manuscript has examined the landscape from multiple perspectives, highlighting both achievements and persistent gaps. The field has evolved from early rule-based and gloss-dependent systems toward end-to-end SLT and SLP pipelines. Recent work has demonstrated the effectiveness of pose-based representations, transformer architectures, and self-supervised learning. Large language models have been successfully integrated into SLT pipelines, improving fluency and semantic coherence. For SLP, advances in neural rendering and diffusion models have enabled higher-fidelity generation. These developments reflect both architectural innovation and the growing availability of large-scale datasets from broadcast and web sources.

Despite this growth, a disconnect remains between what systems achieve on benchmarks and what they deliver in practice. While dataset scale has increased, most large-scale data consists of broadcast interpretations rather than conversational signing, and annotation remains costly. The metrics used to measure progress tell us whether output matches a reference, but not whether it would be understood or accepted by a deaf user. 
These gaps point toward where future effort is needed. Model architectures must better account for the linguistic properties, particularly simultaneity, spatial grammar, and non-manual marking, which remain poorly modeled or ignored entirely. Reducing dependence on expensive gloss annotations through self-supervised and weakly-supervised methods would enable learning from larger and more diverse data sources. Evaluation frameworks should better capture translation quality, potentially through human evaluation protocols or learned metrics that account for sign language structure. Investigation of domain adaptation and few-shot learning are needed to improve generalization across signers and contexts.  

None of this can happen in disciplinary isolation. Modeling decisions benefit from linguistic insight; evaluation gains validity when deaf signers assess output quality; data collection becomes more ethical when communities shape governance from the outset. The technical challenges are not purely engineering problems awaiting algorithmic solutions; they are sociotechnical challenges whose resolution requires sustained collaboration across computer vision, linguistics, and deaf communities. Systems that achieve high scores but produce signing that deaf users find unnatural or incomprehensible have not solved the sign language machine translation problem--they have merely optimized for the wrong objective. A community-informed approach offers not only ethical grounding but methodological correction: by centering the perspectives of those who will use these systems, researchers can better identify which problems matter and what success actually looks like.